\PassOptionsToPackage{warn}{textcomp}
\documentclass[a4paper,9pt]{article}
\usepackage{INTERSPEECH2020}
\usepackage[utf8]{inputenc}
\pdfoutput=1 

\usepackage{microtype}
\usepackage{booktabs}
\usepackage{graphicx}
\usepackage{multirow,array}
\usepackage{tipa}
\usepackage{xcolor,colortbl,soul}
\usepackage{placeins}
\usepackage[hang,flushmargin]{footmisc}
\usepackage{float}

\usepackage{hyperref}

\AtBeginDocument{}%
\AtBeginDocument{}%
\AtBeginDocument{}%

\makeatletter
\newcommand\thefontsize[1]{{#1 The current font size is: \f@size pt\par}}
\makeatother

\let\OLDthebibliography\thebibliography
\renewcommand\thebibliography[1]{
  \OLDthebibliography{#1}
  \setlength{\parskip}{0pt}
  \setlength{\itemsep}{1.97pt plus 1ex}
}

\definecolor{lightgray}{HTML}{EFEFEF}
\newcolumntype{g}{>{\columncolor{lightgray}}c}
\AtBeginDocument{\setlength{\cmidrulekern}{0.3em}}
\usepackage{xspace}

\newcommand{\mtedx}{mTEDx\xspace{}}

\usepackage[disable]{todonotes} 
\makeatletter
\newcommand*\iftodonotes{\if@todonotes@disabled\expandafter\@secondoftwo\else\expandafter\@firstoftwo\fi}  
\makeatother

\newcommand{\note}[4][]{\todo[author=#2,color=#3,size=\scriptsize,fancyline,caption={},#1]{#4}} 
\newlength{\extramargin}
\usepackage{calc}
\iftodonotes{\usepackage[paperwidth=\paperwidth+\extramargin*2,marginparwidth=\marginparwidth+\extramargin,width=\textwidth,height=\textheight]{geometry}}{}
\newcommand{\liz}[2][]{\note[#1]{liz}{teal!40}{#2}}

\newcommand{\matt}[2][]{\note[#1]{matt}{orange!40}{#2}}

\newcommand{\matteo}[2][]{\note[#1]{matteo}{red!40}{#2}}

\newcommand{\jhu}{\textrm{\normalfont \textdyoghlig}}
\newcommand{\umd}{\textrm{\normalfont m}}
\newcommand{\fbk}{\textrm{\normalfont f}}
\newcommand{\corpusurl}{\url{http://www.openslr.org/100}}

\title{The Multilingual TEDx Corpus for Speech Recognition and Translation}

\name{Elizabeth Salesky$^\jhu$~~Matthew Wiesner$^\jhu$~~Jacob Bremerman$^\umd$~~Roldano Cattoni$^\fbk$~~Matteo Negri$^\fbk$ \\ 
Marco Turchi$^\fbk$~~Douglas W. Oard$^\umd$~~Matt Post$^\jhu$}
\address{
  $^\jhu$Johns Hopkins University \\
  $^\umd$University of Maryland \\
  $^\fbk$Fondazione Bruno Kessler}
\email{\{esalesky,wiesner\}@jhu.edu}

\begin{document}
\maketitle 
\begin{abstract}
We present the Multilingual TEDx corpus, built to support speech recognition (ASR) and speech translation (ST) research across many non-English source languages.
The corpus is a collection of audio recordings from TEDx talks in 8 source languages. 
We segment transcripts into sentences and align them to the source-language audio and target-language translations.
The corpus is released along with open-sourced code enabling extension to new talks and languages as they become available. 
Our corpus creation methodology can be applied to more languages than previous work, and creates multi-way parallel evaluation sets.
We provide baselines in multiple ASR and ST settings, including multilingual models to improve translation performance for low-resource language pairs.\liz{sentalert}
\end{abstract}

\section{Introduction}

In speech translation (ST), direct ``end-to-end'' modeling \cite{di2019adapting,inaguma2019multilingual,sperber2019attention,li2021multilingual} replaces the separate automatic speech recognition (ASR) and machine translation (MT) models used in cascaded approaches \cite{stentifordsteer1988,waibel1991} with a single model
that directly translates from source language speech into target language text.
However, this approach requires special parallel resources, which exist only for a few languages and domains.  
We extend the set of publicly available resources with the Multilingual TEDx corpus for ASR and ST.

Most publicly available ST datasets are limited in their language coverage.
The largest ST corpora contain English speech only (\textit{MuST-C}: \cite{di-gangi-etal-2019-mustc,Cattoni2020mustc-v2}, \textit{Librispeech}: \cite{kocabiyikoglu2018augmenting}), while the most widely used corpus with non-English speech (\textit{Fisher-Callhome}: \cite{post2013fisher}) is available only under a paid license. Prior ST corpora relied on third-party forced-aligners, which limited their extensibility and language coverage \cite{di-gangi-etal-2019-mustc,boito2019mass,iranzo2020europarl}.
The last year has seen the creation of ST corpora for additional languages, though most are limited in size to 20--40 hours of translated speech per language pair (\textit{MaSS}, \textit{Europarl-ST}: \cite{boito2019mass,iranzo2020europarl}) and focus on highly specific domains (parliamentary proceedings and the Bible). 
\textit{CoVoST} \cite{wang-etal-2020-covost,wang2020covost2}, built from the Common Voice project \cite{ardila-etal-2020-common-voice}, is a rapidly growing multilingual corpus with translations into and out of English. 
Collecting data in a variety of genres is critical for building generalizable models and assessing real-world performance of ST systems, but many ST corpora consist only of read speech \cite{kocabiyikoglu2018augmenting,boito2019mass,wang-etal-2020-covost,wang2020covost2}. 
Finally, existing multi-target corpora do not have the same segmentations across language pairs \cite{di-gangi-etal-2019-mustc,boito2019mass,iranzo2020europarl}, which complicates the analysis of multilingual models and language transfer. 
In conclusion, the development of generalizeable ST models requires large, diverse, and easily extensible multilingual corpora.

To meet this demand, we present the Multilingual TEDx corpus (\mtedx), built from TEDx talks in 8 languages with their manual transcriptions, and translations in 6 target languages, shown in \autoref{tab:corpus-stats}. \matteo{I'D LIST THE SRC LANGUAGES SOMEWHERE, AS EARLY AS POSSIBLE, AND MAYBE ALSO IN THE ABSTRACT}
We release all code for reproducibility and to enable extension to new languages or additional data as it becomes available. 
The content (TEDx talks) makes the corpus easily combined with common MT resources \cite{cettolo2012wit3}, and its metadata enables use in other applications such as speech retrieval. 
We additionally provide illustrative experiments demonstrating its use for ASR and ST, including multilingual models to improve the performance in low-resource language settings. 
\matteo{Alternative: Through illustrative experiments, we  demonstrate its use for ASR and ST, including the training of multilingual models to improve performance } 
The Multilingual TEDx corpus is released under a CC BY-NC-ND 4.0 license, and can be freely downloaded at \corpusurl.

\section{Corpus Creation}

\subsection{Data Collection}

The source for this corpus is a multilingual set of TEDx\footnote{\url{www.ted.com/about/programs-initiatives/tedx-program}} talks. 
TEDx events, launched in 2009, are independently hosted TED events. TEDx talks share the same format as TED talks. However, while TED talks are all in English, TEDx talks can be in a variety of languages.
Currently there are over 150,000 TEDx recordings across 100+ languages, and more than 3,000 new recordings are added each year.
These talks are short, prepared speeches with an average length of 10 minutes. 
Volunteers create transcripts and translations according to TED guidelines;\footnote{\url{www.ted.com/participate/translate/guidelines}} however, punctuation and annotation for non-speech events contain some amount of noise, such as variance in spacing between punctuation marks, and use of consecutive punctuation. 

Each TEDx talk is stored as a WAV-file sampled at 44.1 or 48kHz. 
The raw transcripts and translations are in WebVTT\footnote{\url{https://en.wikipedia.org/wiki/WebVTT}} subtitle format.
Available metadata include the source language, title, speaker, duration of the audio, keywords, and a short description of the talk. Some talks are missing transcripts, translations, or some metadata fields. 
In this initial version of the corpus, we target source languages other than English with segmental orthography and at least 15 hours of transcribed speech.
Only talks with a provided source language ID and at least source language transcripts are included to enable alignment. 
Not all talks are translated. We release translation pairs where at least 10 hours of speech are translated.  
While our smaller languages\matteo{I'D MENTION THEM IN PARENTHESIS} have limited training data for ASR or ST, they form standard evaluation sets for comparisons on those languages and language pairs.
Corpus statistics by language are shown in \autoref{tab:corpus-stats}.

\begin{table}[!t]
\aboverulesep=1pt
\belowrulesep=1pt
\extrarowheight=\aboverulesep
\addtolength{\extrarowheight}{\belowrulesep}
\aboverulesep=0pt
\belowrulesep=0pt
\resizebox{\columnwidth}{!}{%
\setlength\tabcolsep{4pt} 
\begin{tabular}{ggg|cccccccccc}
\multicolumn{3}{g|}{\textbf{Source Language}}                          & \multicolumn{10}{c}{\textbf{Target Languages} ~~ \textit{Hours, \# Sents}} \\
\cmidrule(r){1-3} \cmidrule(lr){4-13}
& Hours & Utts & \multicolumn{2}{c}{en} & \multicolumn{2}{c}{es}        & \multicolumn{2}{c}{fr}        & \multicolumn{2}{c}{pt}        & \multicolumn{2}{c}{it}        \\
\cmidrule(lr){2-2} \cmidrule(lr){3-3} \cmidrule(lr){4-5} \cmidrule(lr){6-7} \cmidrule(lr){8-9} \cmidrule(lr){10-11} \cmidrule(lr){12-13} 
es         & 189  & \textit{107k}  & 69 & \textit{39k}       & \multicolumn{2}{>{\columncolor{lightgray}[0pt]}c}{  } & 11 & \textit{~~7k}                 & 42 & \textit{24k}                  &  11 & \textit{6k}  \\
fr         & 189  & \textit{119k}  & 50 & \textit{33k}       & 38 & \textit{24k}                  & \multicolumn{2}{>{\columncolor{lightgray}[0pt]}c}{  } & 25 & \textit{16k}            & \multicolumn{2}{c}{---}   \\
pt         & 164 & \textit{~~93k}  & 59 & \textit{34k}       & 26 & \textit{15k}                  & \multicolumn{2}{c}{---}   & \multicolumn{2}{>{\columncolor{lightgray}[0pt]}c}{  } & \multicolumn{2}{c}{---}   \\
it         & 107 & \textit{~~53k}  & 53 & \textit{27k}       & 11 & \textit{~~5k}                 & \multicolumn{2}{c}{---}   & \multicolumn{2}{c}{---}   & \multicolumn{2}{>{\columncolor{lightgray}[0pt]}c}{  } \\
ru         & ~~53 & \textit{~~31k}      & 11 & \textit{~~7k}      & \multicolumn{2}{c}{---}   & \multicolumn{2}{c}{---}   & \multicolumn{2}{c}{---}   & \multicolumn{2}{c}{---}   \\
el         & ~~30 & \textit{~~15k}      & 12 & \textit{~~6k}      & \multicolumn{2}{c}{---}   & \multicolumn{2}{c}{---}   & \multicolumn{2}{c}{---}   & \multicolumn{2}{c}{---}   \\
ar         & ~~18 & \textit{~~14k}      & \multicolumn{2}{c}{---} & \multicolumn{2}{c}{---}   & \multicolumn{2}{c}{---} & \multicolumn{2}{c}{---}   & \multicolumn{2}{c}{---} \\
de        & ~~15 & \textit{~~~~9k}       & \multicolumn{2}{c}{---} & \multicolumn{2}{c}{---}   & \multicolumn{2}{c}{---} & \multicolumn{2}{c}{---}   & \multicolumn{2}{c}{---} \\
\bottomrule
\end{tabular}%
}
\caption{Corpus statistics showing total number of hours and utterances for ASR and MT/ST tasks. Languages are represented by their ISO 639-1 code. Not all talks for a given source language are translated; columns to the right represent the subset of ASR talks available for MT and ST for a given source language.}
\label{tab:corpus-stats}
\end{table}

\subsection{Sentence Alignment for Text}

TEDx subtitles are segments of up to 42 characters per line to optimize screen display, and may not correspond to sentence boundaries.  
We therefore need to create sentence-level segments, which then need to be aligned between source and target languages.
In settings where audio is read from individual sentences, such sentence-level segments are naturally available \cite{wang-etal-2020-covost}; however, when creating a corpus from transcribed and translated speech, these boundaries must be detected, and may differ in number between source and target languages \cite{di-gangi-etal-2019-mustc,Cattoni2020mustc-v2, iranzo2020europarl} depending on translator guidelines.
By keeping consecutive segments from talks in context, we enable context-aware methods. 
The MuST-C and Europarl-ST corpora were created by first splitting the (translated) subtitle text using punctuation, and then independently aligning each of them with the source language transcript using the Gargantua sentence alignment tool \cite{braune-fraser-2010-gargantua}. Segments were allowed to be merged in both the source and target language to create better sentence alignments for a given language pair. 
However, this means that the same talks may have different segmentations in different language pairs, complicating analysis of multilingual models and language transfer.

We took a different approach, designed to produce a multi-way parallel alignment across all available translations for a given source-language transcript.
To do so, we started with a source language segmentation which we then held constant.
We used Punkt \cite{kiss2006unsupervised-punkt}, an unsupervised sentence boundary detector, to generate sentence-level segments for the source language from VTT subtitles.  
Then, we used Vecalign \cite{thompson-koehn-2019-vecalign} to create possible target-language sentences and perform alignment.\footnote{\url{https://github.com/esalesky/vecalign}}
Vecalign uses a dynamic program to compare sentence embeddings, here from LASER \cite{artetxe-schwenk-2019-laser}, to align source and target language sentences. 
To create a space of possible target-language sentences from the target language subtitle segments, we create overlapping windows of 1--20 subtitle concatenations. 
We disallow this behavior on the source side, maintaining parallel Punkt-generated sentences across ASR and MT data for the same source talk. 
We modify the deletion penalty and set the window size such that no source segments are null-aligned. 
\begin{table}[!t]
\centering
\resizebox{\columnwidth}{!}{%
\begin{tabular}{c|cccccccc}
      & \multicolumn{8}{c}{\textbf{Target Languages}} \\
\textbf{Source}  & \multicolumn{2}{c}{en} & \multicolumn{2}{c}{es} & \multicolumn{2}{c}{fr} & \multicolumn{2}{c}{pt} \\
\cmidrule(r){1-1} \cmidrule(lr){2-3} \cmidrule(lr){4-5} \cmidrule(lr){6-7} \cmidrule(lr){8-9} 
 & \textit{V} & \textit{G} &  \textit{V} & \textit{G} &  \textit{V} & \textit{G} &  \textit{V} & \textit{G} \\ 
es     & \bf 26.0 & 24.0   & \multicolumn{2}{>{\columncolor{lightgray}[0pt]}c}{  }   & \bf 1.7 & 1.1     & \bf 39.6 & 36.6  \\
fr     & \bf 28.4 & 27.6   & \bf 30.7  & 27.7     & \multicolumn{2}{>{\columncolor{lightgray}[0pt]}c}{  } & \bf 19.2 & 10.6  \\
pt     & \bf 28.1 & 27.1   & \bf 30.0  & 29.9        & \multicolumn{2}{c}{----}  & \multicolumn{2}{>{\columncolor{lightgray}[0pt]}c}{  } \\
it     & \bf 18.9 & 12.7   & \bf ~~0.7 & ~~0.4       & \multicolumn{2}{c}{----}  & \multicolumn{2}{c}{----}   \\
\bottomrule
\end{tabular}}%
\caption{Comparing sentence alignment methods via downstream MT performance: showing test BLEU with Vecalign sentence alignments (V) and Gargantua sentence alignments (G). }
\label{tab:text-alis}
\end{table}
Relying on punctuation and boundary-indicative text to detect sentences boundaries will not work in all cases; we filter 13 talks which had no strong punctuation, resulting in only one source segment per talk.
Manual inspection revealed these 13 talks contained only singing.

Vecalign has previously been shown to have better intrinsic 
alignment performance than Gargantua \cite{thompson-koehn-2019-vecalign}. 
We lack the ground-truth sentence boundaries and alignments to do intrinsic evaluation here, and so use a downstream application to validate our sentence alignment approach. 
We first create a fixed segmentation of the reference sentences by concatenating all references and resegmenting them with Punkt. 
Next, we independently align the outputs from the Gargantua and Vecalign systems
to these references with minimum edit distance, using a reimplementation\footnote{\url{https://github.com/jniehues-kit/sacrebleu}} of \textsc{mwerSegmenter} \cite{matusov2005mwerSegmenter}. \matt{Let's merge this before the final copy?}\liz{that would be amazing! def possible before camera-ready; right now it's verryy slow--i think he was aiming to improve runtime before submitting a pull request}
We measure MT performance with \textsc{SacreBLEU} \cite{post-2018-sacrebleu} in \autoref{tab:text-alis}.
With the alignments obtained from Vecalign, we see improvements for all language pairs, in some cases slight (\textless1 BLEU) and in others substantial (\textgreater5 BLEU). 
Typically, our approach generates shorter segments, as Gargantua allows segment merges to optimize alignment scores, which may result in more than one sentence per segment. 
On average, Vecalign generates 5k more aligned segments than Gargantua from the same data. 
While this may facilitate training of translation models, it can also cause errors to have proportionally higher impact on BLEU. 
These results suggest our method is as good as the prior method, and preferable for some language pairs (which may have e.g., differing punctuation standards that affect sentence boundary detection and alignment).

\subsection{Sentence Alignment for Speech}
\label{sec:speech-alis}

To align speech segments to the sentence-level text described in the previous section, we need source language acoustic models. We train our own acoustic models for forced alignment and use a sentence-driven audio segmentation as opposed to a speech activity based segmentation.\footnote{\url{https://github.com/m-wiesner/tedx}}

We train a speaker-adapted HMM-GMM ASR system using the audio data and the source language VTT subtitles in \textsc{kaldi} \cite{povey2011kaldi} using MFCC features. During training, the vocabulary is the set of all words in the training transcripts. 
For alignment, we extend the vocabulary to include words that appear in all transcripts. 
We use phonemic lexicons for all languages as provided by \textsc{Wikipron} \cite{wikipron2020lee}. Missing pronunciations are obtained via a grapheme-to-phoneme conversion system trained on the Wikipron seed lexicons using \textsc{Phonetisaurus} \cite{novak2016phonetisaurus}. The IPA phonemes are converted to their equivalent X-SAMPA \cite{wells1995xsampa} ASCII representations using the IPA to X-SAMPA tool in \textsc{Epitran} \cite{epitran2018mortensen} for easier integration with \textsc{kaldi} tools.

Once trained, the acoustic models can be re-used to force-align the segment-level source language text transcripts to the audio. 
From these alignments, we generate a CTM file. 
Each row of the CTM file corresponds to a word from the segment-level transcript, and indicates the audio file in which that word appeared along with the start time and duration of the word according to the forced alignment. 
To obtain time boundaries in the audio data for each sentence, we align the produced CTM file to the source language sentences. 
Roughly speaking, the start time and end times that aligned to the first and last words in a sentence form the sentence time boundaries.

Occasionally, we encounter untranslated portions of the CTM and we would ultimately like to ignore these segments. Likewise the CTM can also have holes relative to the source language sentences. These are portions of the audio data for which we have translations, but no aligned speech. This especially happens when foreign language speech (relative to the source language), non-speech noises, on-screen text or speaker labels are improperly annotated and, as a result, there exists no good alignment between the audio and the reference transcript. Unfortunately, alignment of audio to \emph{properly} annotated text can also occasionally fail. 
Determining which of the two causes was responsible for failed alignments of segments in each talk is not feasible. Instead, we report which segments failed to align so that users can optionally exclude these examples; we note that this occurred in only $\sim$0.1\% of segments. 
We exclude the 2 talks where the majority of segments failed to align, typically signalling a significant transcript mismatch; upon inspection in these cases, the `transcripts' were in fact dialectal translations.

When aligning the CTM files to the reference sentences, we account for these ``holes'' by using a slightly modified Viterbi algorithm: we insert special sentence-boundary tokens (1 per sentence and a start of sentence symbol) at the end of each reference sentence and adjust the insertion and deletion costs to encourage extra speech in the CTM to align to these boundary tokens, as well as to ensure that the boundary tokens are otherwise deleted.
During the back-trace, the sentence time boundaries are updated on non-boundary tokens from the reference by taking the time information of the word in the CTM file aligned to it.
To verify that the acoustic models were successfully trained, we decode held-out, whole audio files for $\sim$10\% of the data in hours of transcribed speech using a 3-gram language model.

\subsection{Standardized Splits}

We created standardized data splits for training, validation, and evaluation by selecting talks which have the highest overlap between target languages and holding out full talks for validation and evaluation until we have $\sim$1,000 utterances ($\sim$2 hours) for each of these splits. 
We use the same talks for both ASR and MT evaluation to ensure there is no contamination between the training and evaluation data between the two tasks in ST.
\liz{note: es-it, fr-ar don't have enough overlapping data to have parallel eval sets with the other es-X, fr-X datasets right now. not releasing fr-ar for now (could add for interspeech); es-it tbd}

\section{Experiments}

\subsection{Automatic Speech Recognition (ASR)}

We created ASR baselines for the corpus using \textsc{kaldi} \cite{povey2011kaldi}. 
Specifically, we trained hybrid models using a CNN-TDNNf neural architecture under the lattice-free maximum mutual information  (LF-MMI) objective \cite{povey2016lfmmi}.
We used exactly the parameters in the \textsc{kaldi} WSJ recipe, except we removed i-vectors from training. 
We resampled audio files to 16kHz and remixed multi-channel talks to one channel. 
The input features were 40-dimensional MFCC features with cepstral mean normalization. 
We decoded with a 4-gram language model trained using \textsc{srilm} \cite{stolcke2002srilm}. 
We used the same pronunciation lexicons described in \autoref{sec:speech-alis}.  
We compared to seq2seq Transformer \cite{vaswani2017attention} models trained in \textsc{fairseq} \cite{ott2019fairseq,wang2020fairseqs2t}, described in \autoref{sec:st-e2e}.
Performance for each language is shown in \autoref{tab:asr-results}. 

\begin{table}[!t]
\centering
\resizebox{\columnwidth}{!}{%
\setlength\tabcolsep{11pt} 
\begin{tabular}{lrrrr}
\toprule
\textbf{Model} & \multicolumn{1}{c}{\textbf{es}} & \multicolumn{1}{c}{\textbf{fr}} & \multicolumn{1}{c}{\textbf{pt}} & \multicolumn{1}{c}{\textbf{it}} \\ 
\midrule
Hybrid LF-MMI  & 16.2 & 19.4 & 20.2 & 16.4 \\ 
Transformer   & 46.4 & 45.6 & 54.8 & 48.0 \\ 
\midrule
\textbf{Model} & \multicolumn{1}{c}{\textbf{ru}} & \multicolumn{1}{c}{\textbf{el}} & \multicolumn{1}{c}{\textbf{ar}} & \multicolumn{1}{c}{\textbf{de}} \\
\midrule
Hybrid LF-MMI  & 28.4 & 25.0 & 80.8 & 42.3      \\
Transformer   & 74.7 & 109.5 & 104.4 & 111.1 \\ 
\bottomrule
\end{tabular}}%
\caption{ASR results on test in WER$\downarrow$. Comparing \textsc{kaldi} Hybrid LF-MMI and \textsc{fairseq} Transformer models (no pretraining).}
\label{tab:asr-results}
\end{table}

The Hybrid LF-MMI models seem more robust to lower-resource conditions. 
In general, our models' performance scaled with data size; however, there were exceptions, such as Italian and Arabic. 
We suspect the poor results in Arabic were due to `transcripts' written in Modern Standard Arabic rather than the significantly different spoken dialect.
Our end-to-end Transformer models did not perform competitively to the hybrid models with or without pretraining; we saw considerable improvements in reducing network size from \cite{wang2020covost2,inaguma2019multilingual}. 
However, they nonetheless enabled considerable improvements for ST when used for encoder initialization. 
When pretrained with larger English datasets from MuST-C, we saw improvements of 15-20 WER for all but our two lowest-resource languages.

\subsection{Machine Translation (MT)}
\label{sec:mt}

\begin{table} [!b]
\setlength\abovecaptionskip{0.0\baselineskip}
\aboverulesep=1pt
\belowrulesep=1pt
\extrarowheight=\aboverulesep
\addtolength{\extrarowheight}{\belowrulesep}
\aboverulesep=0pt
\belowrulesep=0pt
\centering
\resizebox{\columnwidth}{!}{%
\setlength\tabcolsep{8pt} 
\begin{tabular}{ccccg}
\toprule
\multicolumn{1}{c}{\textbf{Src}} & \multicolumn{1}{c}{\textbf{Tgt}} & \multicolumn{1}{c}{\textbf{Individual}} & \multicolumn{1}{c}{\textbf{Multilingual}} & \multicolumn{1}{g}{\textbf{M2M\_100}} \\ 
\cmidrule(r){1-1} \cmidrule(lr){2-2} \cmidrule(lr){3-3} \cmidrule(lr){4-4} \cmidrule(lr){5-5} 
es & en & 25.5  & 24.6 & 34.0 \\
   & pt & 39.3  & 37.3 & 45.3 \\
   & fr & ~~2.0 & 18.1 & 35.5 \\
\cmidrule(lr){1-5}
fr & en & 28.3  & 28.2 & 40.9 \\
   & es & 30.5  & 32.1 & 42.4 \\
   & pt & 19.0  & 30.6 & 35.9 \\
\cmidrule(lr){1-5}
pt & en & 27.9  & 28.8 & 38.7 \\
   & es & 29.9  & 38.4 & 45.8 \\
\cmidrule(lr){1-5}
it & en & 18.9  & 20.9 & 34.6 \\
   & es & ~~1.0 & 25.1 & 44.2 \\
\cmidrule(lr){1-5}
ru & en & ~~1.7 & ~~5.7 & 23.6 \\
\cmidrule(lr){1-5}
el & en & ~~1.9 & ~~7.6 & 29.0 \\
\bottomrule
\end{tabular}}%
\caption{MT results on test in BLEU$\uparrow$. Comparing individual and multilingual MT models trained on this corpus, and performance with a pretrained 1.2B parameter multilingual model (M2M\_100)}
\label{tab:mt-results}
\end{table}

We created text-to-text machine translation baselines using \textsc{fairseq} \cite{ott2019fairseq}. 
We followed the recommended Transformer \cite{vaswani2017attention} hyperparameters for the IWSLT'14 \cite{iwslt2014overview} de-en TED task with a few modifications; 
we used source and target vocabularies of 1k BPE \cite{kudo-richardson-2018-sentencepiece} each for all language pairs, which was previously shown to perform better than larger vocabularies for es-en ST tasks \cite{salesky-etal-2019-exploring}. 
We also used a larger batch size of 16k tokens, which experimentally performed better than batches of 4k or 8k tokens. 
We computed cased detokenized BLEU scores using \textsc{SacreBLEU} \cite{post-2018-sacrebleu}. 
Different community standards for punctuation and case exist between ASR, MT, and ST; while we removed punctuation and lowercased text for ASR evaluation, we did not for translation tasks. 

Baseline results for individual language pairs are shown in \autoref{tab:mt-results}. 
Despite the low-resource settings for all language pairs (4-40k sentences) compared to typical machine translation datasets (100k-10M sentences), we find strong translation performance between more related source and target languages (e.g., es-pt) and language pairs with \textgreater10k sentences. 
Language relatedness here implies a more one-to-one and monotonic alignment between source and target language tokens, which we see yields substantially better translation performance.  

\begin{table*}[!t]
\centering
\resizebox{\textwidth}{!}{%
\setlength\tabcolsep{6pt} 
\begin{tabular}{lrrrrrrrrrrrr}
\toprule
\textbf{Model} & \textbf{es-en} & \textbf{es-pt} & \textbf{es-fr} & \textbf{fr-en} & \textbf{fr-es} & \textbf{fr-pt} & \textbf{pt-en} & \textbf{pt-es} & \textbf{it-en} & \textbf{it-es} & \textbf{ru-en} & \textbf{el-en} \\ 
\cmidrule(lr){1-1} \cmidrule(lr){2-2} \cmidrule(lr){3-3} \cmidrule(lr){4-4} \cmidrule(lr){5-5} \cmidrule(lr){6-6} \cmidrule(lr){7-7} \cmidrule(lr){8-8} \cmidrule(lr){9-9} \cmidrule(lr){10-10} \cmidrule(lr){11-11} \cmidrule(lr){12-12} \cmidrule(lr){13-13}
Bilingual Cascades               & 15.5 & 23.3 & 1.3 & 17.2 & 17.8 & 12.2 & 16.1 & 16.7 & 11.8 & 0.6 & 1.1 & 1.3 \\
Cascades with Multilingual MT    & 15.9 & 23.1 & 11.3 & 18.1 & 19.4 & 19.7 & 17.5 & 22.3 & 13.6 & 15.5 & 3.6 & 5.7 \\
Cascades with M2M\_100           & 21.5 & 26.5 & 23.4 & 25.3 & 26.9 & 23.3 & 22.3 & 26.3 & 21.9 & 28.4 & 13.5 & 17.3 \\
Bilingual End-to-End ST          & 7.0 & 12.2 & 1.7 & 8.9 & 10.6 & 7.9 & 8.1 & 8.7 & 6.4 & 1.0 & 0.7 & 0.6 \\ 
Multilingual End-to-End ST       & 12.3 & 17.4 & 6.1 & 12.0 & 13.6 & 13.2 & 12.0 & 13.7 & 10.7 & 13.1 & 0.6 & 0.8 \\ 
\bottomrule
\end{tabular}%
}
\caption{Speech translation results on test in BLEU$\uparrow$}
\label{tab:st-results}
\end{table*}

\subsubsection{Multilingual Models}

We explored multilingual settings to improve MT performance, in particular, for our lower-resource pairs, e.g., ru-en, el-en, es-fr, and it-es. 
We first trained a single multilingual model for all language pairs in our dataset following the recommended \textsc{fairseq} parameters for the IWSLT'17 multilingual task \cite{iwslt2017overview}. 
This model uses a shared BPE vocabulary of 16k learned jointly across all languages. 
We appended language ID tags \cite{ha2016multi} to the beginning of each sentence for both the encoder and decoder. 
Results are shown compared to individual models for each language pair in \autoref{tab:mt-results}.
We saw clear improvements for our lower-resource pairs --- even though ru-en and el-en share only a target language with the other language pairs, there is clear benefit from additional English target data; for e.g., es-fr and it-es where there is limited parallel data ($\sim$5k sentences) but both the target and source are observed in other language pairs, transfer enables improvements of up to 24 BLEU.  

We additionally demonstrate the performance of large pretrained models on our corpus: 
using the 1.2 billion parameter multilingual M2M\_100 model \cite{fan2020beyond}, trained on 7.5B sentences \cite{schwenk2019ccmatrix,el2019massive} for 100 languages, to directly generate translations for our evaluation sets, we see further improvements. 
While such large MT resources do not have aligned speech, we show in \autoref{sec:st-cascade} they can be combined with smaller speech-aligned ST datasets such as this corpus to enable improved ST performance.

\subsection{Speech Translation (ST)}

Two model paradigms exist for speech translation models: cascades and end-to-end models. 
While cascade models currently represent the state-of-the-art  \cite{iwslt2020findings}, particularly for low-resource language pairs \cite{salesky2020phone}, end-to-end models are closing the performance gap for high-resource settings, as shown in the trends of recent IWSLT evaluation campaigns \cite{iwslt2020findings,iwslt2019findings}. 

We compare both paradigms to provide strong and illustrative baselines for the Multilingual TEDx corpus. 
In addition to cascaded and end-to-end models for individual language pairs, we show how multilingual MT and ST models may be used to improve the performance of the lower-resource language pairs. 
We hope that the presence of larger ASR data as well as both more closely and distantly related languages will provide a realistic sandbox for developing generalizeable models for ST.

\subsubsection{Cascaded Models}
\label{sec:st-cascade}

We trained cascaded ST models for each language pair by combining our best \textsc{kaldi} ASR models with MT models trained on each individual language pair, as well as with the multilingual MT models trained on all available language pairs. 
We used the 1-best ASR transcript after decoding with beam=$10$ as input to the downstream MT model. 
Results are shown in \autoref{tab:st-results}. 
Cascaded ST performance appears to relate more strongly to MT BLEU than ASR WER; see e.g., es-$\forall$ and fr-$\forall$.
The cascade with multilingual MT provided further improvements for all but es-pt, with substantial $\Delta$BLEU for the smallest Romance languages. 

While the majority of machine translation corpora do not have aligned speech, additional text-only data and pretrained models can nonetheless significantly impact ST performance \cite{li2021multilingual,pino2019harnessing}. 
The cascaded ST with M2M\_100 provided multiple BLEU improvements for all pairs, particularly those less-represented in our corpus, though the improvements over our multilingual model are slightly reduced when translating ASR output rather than gold transcripts. 

\vspace{-1.5mm}
\subsubsection{End-to-End Models}
\label{sec:st-e2e}
\vspace{-1.5mm}

End-to-end models for ST, first explored \cite{duong-etal-2016-attentional,berard2016listen,weiss2017sequence} shortly after similar models achieved success for machine translation, have rapidly become the predominantly researched methodology within the field. 
Initial models made use of the pyramidal CNN-LSTM encoders from ASR \cite{zhang2017very}, with recent work adapting VGG \cite{pino2019harnessing} and Transformer models \cite{di2019adapting,inaguma2019multilingual,pham2020relative} for ST. 
We used \textsc{fairseq} \cite{wang2020fairseqs2t} to train end-to-end Transformer models for ST, using 80-dimensional log mel filterbank features with cmvn and SpecAugment \cite{park2019specaugment}, and 1-d convolutions for downsampling of speech in time.
We used a reduced network depth of 6 encoder layers and 3 decoder layers for bilingual experiments, and $2\times$ larger for multilingual experiments.
We pretrained encoders for ST on the ASR task with the full ASR data for a given source language using the same Transformer architecture and objective, shown to improve the performance of end-to-end ST models particularly for lower-resource language pairs \cite{bansal-etal-2019-pre}. 
We used BPE vocabularies of 1k for ASR and bilingual ST. 
We averaged the last 10 checkpoints for inference. 
Results are shown in \autoref{tab:st-results}. 
At these data sizes, cascaded model performance was stronger. 
Performance trends across language pairs is similar for bilingual end-to-end and cascaded models. 

Multilingual models for ST have recently been explored by combining multiple corpora \cite{inaguma2019multilingual} or adapting pretrained wav2vec and mBART models to the ST task with CoVoST \cite{li2021multilingual}. 
Here we train an end-to-end multilingual model for all language pairs in the Multilingual TEDx corpus, with a joint vocabulary of 8k BPE. 
We pretrain the encoder with Spanish ASR. 
We see improvements for all languages over bilingual end-to-end models and further over the lowest-resource bilingual cascades, but a gap still remains between the multilingual end-to-end model and our multilingual cascades.

\section{Conclusion}
\vspace{-1mm}

We release the Multilingual TEDx corpus to facilitate speech recognition and speech translation research across more languages, and enable comparison between a broad range of existing methodologies such as cascaded and end-to-end ST models and multilingual and transfer learning techniques. 
The talk format of the corpus potentially also enables additional applications such as speech retrieval. 
We make our data and corpus creation methodology publicly available via a CC BY-NC-ND 4.0 license.

\section{Acknowledgments}
\vspace{-1mm}

We thank Jan Niehues, Brian Thompson, and Rachel Wicks.

\bibliographystyle{IEEEtran}
{\footnotesize
\bibliography{bibliography}}

\begin{thebibliography}{10}
\providecommand{\url}[1]{#1}
\csname url@samestyle\endcsname
\providecommand{\newblock}{\relax}
\providecommand{\bibinfo}[2]{#2}
\providecommand{\BIBentrySTDinterwordspacing}{\spaceskip=0pt\relax}
\providecommand{\BIBentryALTinterwordstretchfactor}{4}
\providecommand{\BIBentryALTinterwordspacing}{\spaceskip=\fontdimen2\font plus
\BIBentryALTinterwordstretchfactor\fontdimen3\font minus
  \fontdimen4\font\relax}
\providecommand{\BIBforeignlanguage}[2]{{%
\expandafter\ifx\csname l@#1\endcsname\relax
\typeout{** WARNING: IEEEtran.bst: No hyphenation pattern has been}%
\typeout{** loaded for the language `#1'. Using the pattern for}%
\typeout{** the default language instead.}%
\else
\language=\csname l@#1\endcsname
\fi
#2}}
\providecommand{\BIBdecl}{\relax}
\BIBdecl

\bibitem{di2019adapting}
M.~A. Di~Gangi, M.~Negri, and M.~Turchi, ``Adapting transformer to end-to-end
  spoken language translation,'' in \emph{Proc. of Interspeech}, 2019.

\bibitem{inaguma2019multilingual}
H.~Inaguma, K.~Duh, T.~Kawahara, and S.~Watanabe, ``Multilingual end-to-end
  speech translation,'' in \emph{Proc. of ASRU}, 2019.

\bibitem{sperber2019attention}
M.~Sperber, G.~Neubig, J.~Niehues, and A.~Waibel, ``Attention-passing models
  for robust and data-efficient end-to-end speech translation,''
  \emph{Transactions of the Association for Computational Linguistics}, 2019.

\bibitem{li2021multilingual}
X.~Li, C.~Wang, Y.~Tang \emph{et~al.}, ``Multilingual speech translation with
  efficient finetuning of pretrained models,'' 2021, arXiv preprint
  arXiv:2010.12829.

\bibitem{stentifordsteer1988}
F.~W. Stentiford and M.~G. Steer, ``Machine translation of speech,''
  \emph{British Telecom Technology Journal}, vol.~6, pp. 116--123, 04 1988.

\bibitem{waibel1991}
A.~{Waibel}, A.~N. {Jain}, A.~E. {McNair} \emph{et~al.}, ``Janus: a
  speech-to-speech translation system using connectionist and symbolic
  processing strategies,'' in \emph{Proc. of ICASSP}, 1991.

\bibitem{di-gangi-etal-2019-mustc}
M.~A. Di~Gangi, R.~Cattoni, L.~Bentivogli, M.~Negri, and M.~Turchi,
  ``{M}u{ST}-{C}: a {M}ultilingual {S}peech {T}ranslation {C}orpus,'' in
  \emph{Proc. of NAACL}, 2019.

\bibitem{Cattoni2020mustc-v2}
R.~Cattoni, M.~A. {Di Gangi}, L.~Bentivogli, M.~Negri, and M.~Turchi, ``Must-c:
  A multilingual corpus for end-to-end speech translation,'' \emph{Computer
  Speech \& Language}, vol.~66, p. 101155, 2021.

\bibitem{kocabiyikoglu2018augmenting}
A.~Kocabiyikoglu, L.~Besacier, and O.~Kraif, ``{Augmenting Librispeech with
  French Translations: A Multimodal Corpus for Direct Speech Translation
  Evaluation},'' in \emph{Proc. of LREC}, 2018.

\bibitem{post2013fisher}
M.~Post, G.~Kumar, A.~Lopez, D.~Karakos, C.~Callison-Burch, and S.~Khudanpur,
  ``Improved speech-to-text translation with the {F}isher and {C}allhome
  {S}panish--{E}nglish speech translation corpus,'' 2013.

\bibitem{boito2019mass}
M.~Z. Boito, W.~N. Havard, M.~Garnerin \emph{et~al.}, ``Ma{SS}: A large and
  clean multilingual corpus of sentence-aligned spoken utterances extracted
  from the bible,'' \emph{Proc. of LREC}, 2020.

\bibitem{iranzo2020europarl}
J.~Iranzo-S{\'a}nchez, J.~A. Silvestre-Cerd{\`a}, J.~Jorge \emph{et~al.},
  ``{Europarl-ST}: A multilingual corpus for speech translation of
  parliamentary debates,'' in \emph{Proc. of ICASSP}.\hskip 1em plus 0.5em
  minus 0.4em\relax IEEE, 2020.

\bibitem{wang-etal-2020-covost}
C.~Wang, J.~Pino, A.~Wu, and J.~Gu, ``{C}o{V}o{ST}: A diverse multilingual
  speech-to-text translation corpus,'' in \emph{Proc. of LREC}, 2020.

\bibitem{wang2020covost2}
C.~Wang, A.~Wu, and J.~Pino, ``{C}o{V}o{ST} 2: A massively multilingual
  speech-to-text translation corpus,'' 2020.

\bibitem{ardila-etal-2020-common-voice}
R.~Ardila, M.~Branson \emph{et~al.}, ``Common voice: A massively-multilingual
  speech corpus,'' in \emph{Proc. of LREC}, 2020.

\bibitem{cettolo2012wit3}
M.~Cettolo, C.~Girardi, and M.~Federico, ``Wit3: Web inventory of transcribed
  and translated talks,'' in \emph{Proc. of EAMT}, 2012.

\bibitem{braune-fraser-2010-gargantua}
F.~Braune and A.~Fraser, ``Improved unsupervised sentence alignment for
  symmetrical and asymmetrical parallel corpora,'' in \emph{Proc. of COLING},
  2010.

\bibitem{kiss2006unsupervised-punkt}
T.~Kiss and J.~Strunk, ``Unsupervised multilingual sentence boundary
  detection,'' \emph{Computational Linguistics}, 2006.

\bibitem{thompson-koehn-2019-vecalign}
B.~Thompson and P.~Koehn, ``{V}ecalign: Improved sentence alignment in linear
  time and space,'' in \emph{Proc. of EMNLP}, 2019.

\bibitem{artetxe-schwenk-2019-laser}
M.~Artetxe and H.~Schwenk, ``Massively multilingual sentence embeddings for
  zero-shot cross-lingual transfer and beyond,'' \emph{Transactions of the
  Association for Computational Linguistics}, 2019.

\bibitem{matusov2005mwerSegmenter}
E.~Matusov, G.~Leusch, O.~Bender, and H.~Ney, ``Evaluating machine translation
  output with automatic sentence segmentation,'' in \emph{Proc. of IWSLT},
  2005.

\bibitem{post-2018-sacrebleu}
M.~Post, ``A call for clarity in reporting {BLEU} scores,'' in \emph{Proc. of
  WMT}, 2018.

\bibitem{povey2011kaldi}
D.~Povey, A.~Ghoshal, G.~Boulianne \emph{et~al.}, ``The {K}aldi speech
  recognition toolkit,'' in \emph{Proc. of ASRU}, 2011.

\bibitem{wikipron2020lee}
J.~L. Lee, L.~F. Ashby \emph{et~al.}, ``Massively multilingual pronunciation
  modeling with {W}iki{P}ron,'' in \emph{Proc. of LREC}, 2020.

\bibitem{novak2016phonetisaurus}
J.~R. Novak, N.~Minematsu, and K.~Hirose, ``Phonetisaurus: Exploring
  grapheme-to-phoneme conversion with joint n-gram models in the {WFST}
  framework,'' \emph{Natural Language Engineering}, 2016.

\bibitem{wells1995xsampa}
J.~C. Wells, ``Computer-coding the {IPA}: A proposed extension of {SAMPA},''
  1995/2000.

\bibitem{epitran2018mortensen}
D.~R. Mortensen, S.~Dalmia, and P.~Littell, ``Epitran: Precision {G2P} for many
  languages,'' in \emph{Proc. of LREC}, 2018.

\bibitem{povey2016lfmmi}
D.~Povey, V.~Peddinti, D.~Galvez \emph{et~al.}, ``Purely sequence-trained
  neural networks for {ASR} based on lattice-free {MMI},'' in \emph{Proc. of
  Interspeech}, 2016.

\bibitem{stolcke2002srilm}
A.~Stolcke, ``{SRILM} - an extensible language modeling toolkit,'' in
  \emph{Proc. of ICASSP}, 2002.

\bibitem{vaswani2017attention}
A.~Vaswani, N.~Shazeer, N.~Parmar, J.~Uszkoreit, L.~Jones, A.~N. Gomez,
  {\L}.~Kaiser, and I.~Polosukhin, ``Attention is all you need,'' in
  \emph{Proc. of NeurIPS}, 2017.

\bibitem{ott2019fairseq}
M.~Ott, S.~Edunov, A.~Baevski, A.~Fan, S.~Gross, N.~Ng, D.~Grangier, and
  M.~Auli, ``fairseq: A fast, extensible toolkit for sequence modeling,'' in
  \emph{Proc. of NAACL-HLT: Demonstrations}, 2019.

\bibitem{wang2020fairseqs2t}
C.~Wang, Y.~Tang, X.~Ma, A.~Wu, D.~Okhonko, and J.~Pino, ``fairseq {S2T}: Fast
  speech-to-text modeling with fairseq,'' in \emph{Proc. of AACL:
  Demonstrations}, 2020.

\bibitem{iwslt2014overview}
M.~Cettolo, J.~Niehues, S.~St{\"u}ker, L.~Bentivogli, and M.~Federico, ``Report
  on the 11th iwslt evaluation campaign,'' in \emph{Proc. of IWSLT}, 2014.

\bibitem{kudo-richardson-2018-sentencepiece}
T.~Kudo and J.~Richardson, ``{S}entence{P}iece: A simple and language
  independent subword tokenizer and detokenizer for neural text processing,''
  in \emph{Proc. of EMNLP: Demonstrations}, 2018.

\bibitem{salesky-etal-2019-exploring}
E.~Salesky, M.~Sperber, and A.~W. Black, ``Exploring phoneme-level speech
  representations for end-to-end speech translation,'' in \emph{Proc. of ACL},
  2019.

\bibitem{iwslt2017overview}
M.~Cettolo, M.~Federico \emph{et~al.}, ``Overview of the {IWSLT} 2017
  evaluation campaign,'' in \emph{Proc. of IWSLT}, 2017.

\bibitem{ha2016multi}
T.-L. Ha, J.~Niehues, and A.~Waibel, ``Toward multilingual neural machine
  translation with universal encoder and decoder,'' in \emph{Proc. of IWSLT},
  2016.

\bibitem{fan2020beyond}
A.~Fan, S.~Bhosale, H.~Schwenk \emph{et~al.}, ``Beyond {E}nglish-centric
  multilingual machine translation,'' \emph{arXiv preprint}, 2020.

\bibitem{schwenk2019ccmatrix}
H.~Schwenk, G.~Wenzek, S.~Edunov, E.~Grave, and A.~Joulin, ``Ccmatrix: Mining
  billions of high-quality parallel sentences on the web,'' \emph{arXiv
  preprint arXiv:1911.04944}, 2019.

\bibitem{el2019massive}
A.~El-Kishky, V.~Chaudhary, F.~Guzman, and P.~Koehn, ``A massive collection of
  cross-lingual web-document pairs,'' \emph{arXiv preprint arXiv:1911.06154},
  2019.

\bibitem{iwslt2020findings}
E.~Ansari, N.~Bach, O.~Bojar \emph{et~al.}, ``Findings of the {IWSLT} 2020
  evaluation campaign,'' in \emph{Proc. of IWSLT}, 2020.

\bibitem{salesky2020phone}
E.~Salesky and A.~W. Black, ``Phone features improve speech translation,''
  \emph{Proc. of ACL}, 2020.

\bibitem{iwslt2019findings}
J.~Niehues, R.~Cattoni, S.~St{\"u}ker, M.~Negri, M.~Turchi, E.~Salesky,
  R.~Sanabria, L.~Barrault, L.~Specia, and M.~Federico, ``The {IWSLT} 2019
  evaluation campaign,'' in \emph{Proc. of IWSLT}, 2019.

\bibitem{pino2019harnessing}
J.~Pino, L.~Puzon, J.~Gu, X.~Ma, A.~D. McCarthy, and D.~Gopinath, ``Harnessing
  indirect training data for end-to-end automatic speech translation: Tricks of
  the trade,'' in \emph{Proc. of IWSLT}, 2019.

\bibitem{duong-etal-2016-attentional}
L.~Duong, A.~Anastasopoulos, D.~Chiang, S.~Bird, and T.~Cohn, ``An attentional
  model for speech translation without transcription,'' in \emph{Proc. of
  NAACL}, 2016.

\bibitem{berard2016listen}
A.~B{\'e}rard, O.~Pietquin \emph{et~al.}, ``Listen and translate: A proof of
  concept for end-to-end speech-to-text translation,'' 2016, arXiv preprint
  arXiv:1612.01744.

\bibitem{weiss2017sequence}
R.~J. Weiss, J.~Chorowski, N.~Jaitly, Y.~Wu, and Z.~Chen,
  ``Sequence-to-sequence models can directly translate foreign speech,''
  \emph{Proc. of Interspeech}, 2017.

\bibitem{zhang2017very}
Y.~Zhang, W.~Chan, and N.~Jaitly, ``Very deep convolutional networks for
  end-to-end speech recognition,'' \emph{Proc. of ICASSP}, 2017.

\bibitem{pham2020relative}
N.-Q. Pham, T.-L. Ha, T.-N. Nguyen, T.-S. Nguyen, E.~Salesky, S.~Stueker,
  J.~Niehues, and A.~Waibel, ``Relative positional encoding for speech
  recognition and direct translation,'' 2020.

\bibitem{park2019specaugment}
D.~S. Park, W.~Chan, Y.~Zhang, C.-C. Chiu, B.~Zoph, E.~D. Cubuk, and Q.~V. Le,
  ``Specaugment: A simple data augmentation method for automatic speech
  recognition,'' in \emph{Proc. of Interspeech}, 2019.

\bibitem{bansal-etal-2019-pre}
S.~Bansal, H.~Kamper, K.~Livescu \emph{et~al.}, ``Pre-training on high-resource
  speech recognition improves low-resource speech-to-text translation,'' in
  \emph{Proc. of NAACL}, 2019.

\end{thebibliography}

\end{document}